\documentclass{elsarticle}

\usepackage{hyperref}
\usepackage{placeins}
\usepackage{graphicx}
\usepackage{multirow}
\usepackage{rotating}
\usepackage{nameref}
\usepackage{amssymb, amsfonts, amsmath}
\usepackage{array}
\usepackage{stfloats}
\usepackage{booktabs, tabularx}
\usepackage{subfig}
\usepackage{alphalph}
\usepackage{ragged2e}
\newcolumntype{P}[1]{>{\centering\arraybackslash}p{#1}}

\usepackage{xcolor}
\definecolor{light-gray}{HTML}{E5E4E2}

\usepackage{setspace}
\usepackage{nicefrac}
\usepackage{color}
\usepackage{todonotes}
\usepackage{multirow}

\usepackage{acronym}
\newacro{simon}[SIMON]{SIMilarity-based sentiment projectiON}

\journal{Knowledge-Based Systems}

\begin{document}

\begin{frontmatter}
\title{MoralStrength: Exploiting a Moral Lexicon and Embedding Similarity for Moral Foundations Prediction}

% to be updated:

% \author{Oscar Araque\corref{mycorrespondingauthor}}
% \ead{o.araque@upm.es}
% \cortext[mycorrespondingauthor]{Corresponding author}
% \address{Intelligent Systems Group, Universidad Polit\'ecnica de Madrid, Avenida Complutense, 30, Madrid, Spain}

% \author{Lorenzo Gatti}
% \ead{l.gatti@utwente.nl}
% \address{Human Media Interaction Lab, University of Twente, Enschede, The Netherlands}

% \author{Kyriaki Kalimeri}
% \ead{kkalimeri@acm.org}
% \address{Data Science Laboratory, ISI Foundation, Turin, Italy}

\author[1]{Oscar Araque\corref{mycorrespondingauthor}}
\ead{o.araque@upm.es}

\author[2]{Lorenzo Gatti}
\ead{l.gatti@utwente.nl}

\author[3]{Kyriaki Kalimeri}
\ead{kkalimeri@acm.org}

\cortext[mycorrespondingauthor]{Corresponding author}
\address[1]{Intelligent Systems Group, Universidad Polit\'ecnica de Madrid, Madrid, Spain}
\address[2]{Human Media Interaction Lab, University of Twente, Enschede, The Netherlands}
\address[3]{Data Science \& Digital Philanthropies Laboratory, ISI Foundation, Turin, Italy}

%\maketitle

\begin{abstract}
Moral rhetoric plays a fundamental role in how we perceive and interpret the information we receive, greatly influencing our decision-making process. Especially when it comes to controversial social and political issues,  our opinions and attitudes are hardly ever based on evidence alone. 
The Moral Foundations Dictionary (MFD) was developed to operationalize moral values in the text. 
In this study, we present \textit{MoralStrength}, a lexicon of approximately 1,000 lemmas, obtained as an extension of the Moral Foundations Dictionary, based on WordNet synsets. 
Moreover, for each lemma it provides with a crowdsourced numeric assessment of \textit{Moral Valence}, indicating the strength with which a lemma is expressing the specific value. 
We evaluated the predictive potentials of this moral lexicon, defining three utilization approaches of increased complexity, ranging from lemmas' statistical properties to a deep learning approach of word embeddings based on semantic similarity.
Logistic regression models trained on the features extracted from \textit{MoralStrength}, significantly outperformed the current state-of-the-art, reaching an F1-score of 87.6\% over the previous 62.4\% (p-value$<0.01$), and an average F1-Score of 86.25\% over six different datasets.
Such findings pave the way for further research, allowing for an in-depth understanding of moral narratives in text for a wide range of social issues.
\end{abstract}


\end{frontmatter}

%User-generated content and s
\section{Introduction}

%%Defining and quantifying such complex notions in a reliable way is challenging \cite{Wiebe2005,Kalimeri2019}.
% \cite{Pennebaker2003}.
%Hoover2019: 35k dictionary
%Rezapour2019: new paper with dictionary
Language usage reflects our thoughts, emotions, values, and culture, as we communicate with others.
With the burst of online communication and social media, 
people are empowered to express and broadcast their opinions on contentious issues, timely, and at greater scale.
This unprecedented opportunity allows scientists and policymakers to study phenomena such as opinion formation, radicalization, and polarisation in society, as they happen.
%, having a complementary view of the society.
%we now have access to spontaneous digital behaviors and  user-generated content, .
% 

In this study, we propose a lexicon for detecting and quantifying the moral rhetoric behind people's judgments, as reflected in spontaneous digital interactions.
%from user generated content
Moral values influence the way we rationalize and take a stance upon controversial topics, like abortion~\cite{Sagi2014us}, homosexuality~\cite{Sagi2014us}, climate change~\cite{Wolsko2016},  or even vaccine hesitancy~\cite{Amin2017,Kalimeri2019human}.
They are also closely related to our political views~\cite{Miles2015} and the opinion formation mechanisms regarding immigration~\cite{Grover2019}, political extremism~\cite{Alizadeh2019,Sagi2014us}, and poverty~\cite{Low2015}.
Recently, scientists also showed that moral values could be employed to detect violent protests~\cite{Mooijman2018} based on user-generated text.

We operationalize morality via the Moral Foundations Theory (MFT)~\cite{Graham2009}, which expresses the psychological basis of morality in terms of innate intuitions, defining the following five foundations:
 \emph{care/harm}, % basic concerns for the suffering of others,  including virtues of caring and compassion;
 \emph{fairness/cheating}, % concerns about unfair treatment,  inequality,  and more abstract notions of justice;
 \emph{loyalty/betrayal}, % concerns related to obligations of group membership,  such as loyalty,  self-sacrifice, and vigilance against betrayal;
 \emph{authority/subversion}, % concerns related to social order and the obligations of hierarchical relationships
%such as obedience,  respect, % and proper role fulfilment.
and \emph{purity/degradation} (see \cite{Haidt2007,  Haidt2004}). % concerns about physical and spiritual contagion,  including virtues of chastity,  wholesomeness, and control of desires.
%Moral traits are considered to be a higher level psychological construct with respect to the dispositional traits of personality expressed in the Five-Factor model (\cite{Costa1992}). Dan McAdams's three-level account of personality (dispositional traits,  characteristic adaptations,  and life stories) is clarifying how and when dispositions and attitudes towards interpersonal and intergroup processes relate with persuasion and communication narratives (\cite{McAdams1995,McAdams2006}).
Even if in its infancy, MFT is the most well-established theory in the psychology and social sciences. It is also broadly adopted in the computational social science field since it defines a clear taxonomy of values together with a term dictionary, the Moral Foundations Dictionary (MFD, hereafter) \cite{Graham2009}, which is an essential resource for natural language processing applications.
The creators of the MFD, highlight the difficulty of creating such a resource since linguistic, cultural, and historical context reflect on language usage. 
Among the most significant limitations of the MFD, we have: (i) a limited amount of lemmas and stem of words; (ii) ``radical" lemmas rarely used in everyday language, for instance,  ``homologous'' and ``apostasy''; and (iii) an association with a moral bipolar scale, so-called vice and virtue, but without any indication of ``strength''. 

Here, we address precisely these shortcomings; initially, we expanded the existing MFD using the WordNet lexical database~\cite{Miller1995} and then, we provide a set of normative ratings for empirical assessment of morality.
The resulting lexicon, namely \textit{MoralStrength}, offers approximately three times more lemmas while going beyond the binary nature of the MFD.
Moreover, we present a machine learning framework exploring the potentials of  \textit{MoralStrength} in predicting the moral narratives from the user-generated text.

The suggested framework includes three models of increasing complexity; two of them are based on straightforward feature extraction from lemma frequencies and statistical properties, while the third one is based on embedding representation of semantic similarity. 
%The semantic similarity is a metric computed between the text and the lexicons created ad-hoc, through the exploitation of embedding representations. 
%All models are then employed to predict the moral narratives of a given text.
%In this way, we advance the current concept of the bipolar association of a given word with a moral dimension, to a state where we have not only a richer dictionary, but also an assessment of ``moral valence'' for each lemma.
We thoroughly evaluate the proposed framework employing the Moral Foundations Twitter Corpus (MFTC) \cite{Hoover2019}. 
The MFTC corpus is a collection of seven Twitter dataset previously employed in studies related to moral detection from text. It consists of approximately 35,000 tweets along with their respective annotations according to the MFT foundations regarding critical social issues. 

Importantly, the performance of our approach in predicting morality is outperforming the current state-of-the-art methods.
Our results show that the pure textual representations emerged from the \textit{MoralStrength} lexicon greatly benefit the performance of the prediction. 
These findings pave the way towards a more in-depth understanding of moral judgments, dispositions, and attitudes formation from spontaneous digital data.

Hence, we contribute to the research and policymaking communities with a useful resource and a concrete framework that can be employed for analyzing large scale user-generated communications, or even nowcasting people's attitudes and opinions on controversial phenomena.
When it comes to critical social issues, the proposed approach can provide insights to understand personal narratives and viewpoints better, but also how people will potentially reason on the information they receive.
Such knowledge is essential for policymaking specialists to design effective communication campaigns that appeal to people's values, given the ever-increasing penetration of social media to the population.

\section{Related Literature}

Psychologists and social scientists have systematically analyzed text data to address their research questions. Back in the late '60s, the Harvard General Inquirer dictionary~\cite{stone1966general} was the precursor of sentiment analysis which was to become, together with opinion extraction, a core theme of natural language processing (NLP).
Ever since scientists gradually increased complexity moving from simple techniques (e.g. the unsupervised and partially rule-based approach of~\cite{turney2002thumbs}) to sophisticated methods that try to determine the context of words (e.g. \cite{muhammad2016contextual}). The topics addressed also became more challenging tackling notions such as irony and sarcasm~ \cite{sulis2016figurative}. 
%Iliev et al.~\cite{Iliev2015} present an extensive review of how automated text analysis has been used by social scientists. 
With time, not only the methods became more sophisticated but also the tasks become more ambitious.
%; the research community moved from detection of sentiment~\cite{poria2014emosenticspace} and emotions~\cite{depechemoodpp}.}
Extensive studies on linguistic markers of sentiment and affect~\cite{Strapparava2008,Liu2015,muhammad2016contextual,poria2014emosenticspace,depechemoodpp} paved the way to assess more complex constructs such as  personality~\cite{Schwartz2013,Yarkoni2010} and human values~\cite{Boyd2015,Chen2014}.
Moral values are considered to be a higher level construct with respect to personality traits, determining how and when dispositions and attitudes relate with our life stories and narratives~\cite{Mcadams2006}.
Here we provide a brief literature review of the studies regarding moral values assessment from textual data. As in sentiment and personality analysis, also, in this case, pioneer works followed a dictionary-based approach, while the current state of the art performance is based on deep learning.

%%core topic of this paper, and one of the most recent intersections between NLP and social sciences:

%Iliev et al. \cite{Iliev2015} provided a review of the most popular approaches to automated text analysis from the perspective of social scientists.
%As the role of natural language processing is steadily growing, the methodologies for text analysis for social sciences range from completely controlled, dictionary-based studies~\cite{Boyd2015} to purely data-driven modelling of human values~\cite{Schwartz2013}.
%Here, we provide a brief review of the current literature on moral values assessment from textual data, starting from dictionary based studies to deep learning approaches.

% K: MFD studies with LIWC
The first vocabulary developed to assess the moral values from textual data was the Moral Foundations Dictionary (MFD)~\cite{Graham2009}. It was used together with the Linguistic Inquiry and Word Count (LIWC) software \cite{Tausczik2010} to estimate moral traits and to investigate differences in moral concerns between different cultural groups.
Clifford et al.~\cite{Clifford2013} employed the MFD for performing manual text analysis of 12 years of coverage in the New York Times focusing on political debate in the US.
Teernstra et al.~\cite{Teernstra2016} assessed the political debate regarding the ``Grexit'' from approximately 8,000 tweets. They compared the performance of using the raw data, bi-grams, and the MFD features in employing basic machine learning models, namely, Naive Bayes (NB) and Maximum Entropy (ME). 
They concluded that pure machine learning is preferable to dictionary approaches since it has similar prediction accuracy while using fewer assumptions.
In this study, we follow a similar approach to \cite{Teernstra2016}; however, we propose an expanded version of the MFD, including also the moral valence per lemma. Moreover, we employ logistic regression models to infer moral values from uni-grams combined with lexical features. % check this again.

% K: LDA, LSA staff
Dehghani et al.~\cite{Dehghani2014} examined the differences between liberal and conservative moral value systems using a hierarchical generative topic modelling technique, based on Latent Dirichlet Allocation (LDA)~\cite{Blei2003}, to enable the unsupervised detection of topics in their corpus of liberal and conservative weblogs. They used small sets of words selected from the MFD as seeds to encourage the emergence of issues related to different moral concerns and examined similarities and differences in how such matters are expressed between these groups. Consistently with findings in moral psychology, they demonstrate that there are significant differences in how liberals and conservatives construct their moral belief systems. 
Sagi et al.~\cite{sagi2014measuring} employed the same framework to study moral rhetoric in text for a specific case study, the US Federal shutdown of 2013 \cite{Sagi2014us}, where they examined the role of morals in intra- and inter-community differences of political party retweets.
In both works, they were based on the framework presented in \cite{Dehghani2014}, where LDA was employed to create a co-occurrence matrix on which the similarity between the texts and the vectors representing the different MFT moral traits was computed. % the reduced dimensions with SVD.
In a similar approach, Kaur et al.~\cite{Kaur2016} attempted to quantify the moral loadings of text, based on the Latent Semantic Analysis (LSA). They used a bag-of-words model, representing the entire corpus by a word-context matrix. Then they reduced its dimensionality obtaining low-dimensional word vectors, in which similar vectors represent similar meaning words.
Our study is presenting a different approach since we do not use LSA representations, but rather pre-trained word embeddings models.
Although pre-trained word embeddings do not contain domain-specific knowledge, they express language regularities encoded as offsets in the resulting vector space.
The proposed representations based on the work of Araque et al.~\cite{araque2018similarity}, exploit precisely the similarity between the analyzed text and a selection of words with moral content.

%K: DDR and embeddings
More recently, Garten et al.~\cite{Garten2016} employed the MFD to detect moral rhetoric in general, and more specifically, shifts in long political speeches over time.
Then, based on psychological dictionaries and semantic similarity to quantify the presence of moral sentiment around a given topic, Garten et al.~\cite{Garten2018}, proposed the Distributed Dictionary Representations (DDR) method. 
Showing promising results, DDR was also employed by Hoover et al.~\cite{Hoover2018} to detect moral values in charitable giving. Later on, Garten et al.~\cite{Garten2019} extended the method, incorporating demographic embeddings into the language representations.
Our approach is based on an expanded version of the MFD, with evaluated manual annotations regarding the moral valence of each lemma, that can be incorporated in computational frameworks.

%, thus leveraging this extensive resource.
% The expanded vocabulary has been manually annotated, offering a moral valence that can be used computationally.

% K: neural networks
In a study more similar to ours, attempting to predict moral values involved in Twitter posts, Lin et al. \cite{Lin2018} proposed a method that automatically acquires background knowledge to improve the moral value prediction, pointing out the difficulty of the task also for human experts.
Based on the work of \cite{Lin2018} and \cite{Graham2009}, \cite{Mooijman2018} predicted the moral sentiment of the tweets. 
Their model consists of three layers, an embedding (lookup) layer, a recurrent neural network (RNN) with long short-term memory (LSTM) \cite{Hochreiter1997} and an output layer.
The first layer converts words in an input tweet to a sequence of pre-trained word embeddings, the LSTM layer processes these embeddings and outputs a fixed-sized vector which encodes critical information for moral value prediction, while a vector representing the percentage of words that match each category in the Moral Foundations Dictionary \cite{Graham2009} are concatenated with the LSTM feature vector.
Our approach is differentiating to this one since we employ word embeddings to compute the similarity between words rather than directly feeding them in a neural network architecture.
Along the same line, Rezapour et al. \cite{Rezapour2019} investigated the relationship between moral values and stance towards a series of social issues. Their findings show that enhancing the original MFD improves the prediction accuracy of morality in text. They expanded the original MFD and employed a series of machine learning classifiers (SVM, RF and LSTM) predicting the moral traits.
This study underlines the importance of expanding the MFD; we go one step further introducing the notion of moral ``strength'' while showing how more abundant information is improving the overall accuracy of the models.

The core contribution of this study is the extended lexicon of moral lemmas with the respective moral valence.
To explore the properties and full potentials of the lexicon, we suggest three different models of increasing complexity, demonstrating the value of this resource.
The proposed approaches range from feature engineering methods to a system which employs word embeddings of semantic similarity based on the work of Araque et al. \cite{araque2018similarity}.

\section{Materials and Methods}

\begin{figure}[!ht]
  \centering
    \includegraphics[width=\linewidth]{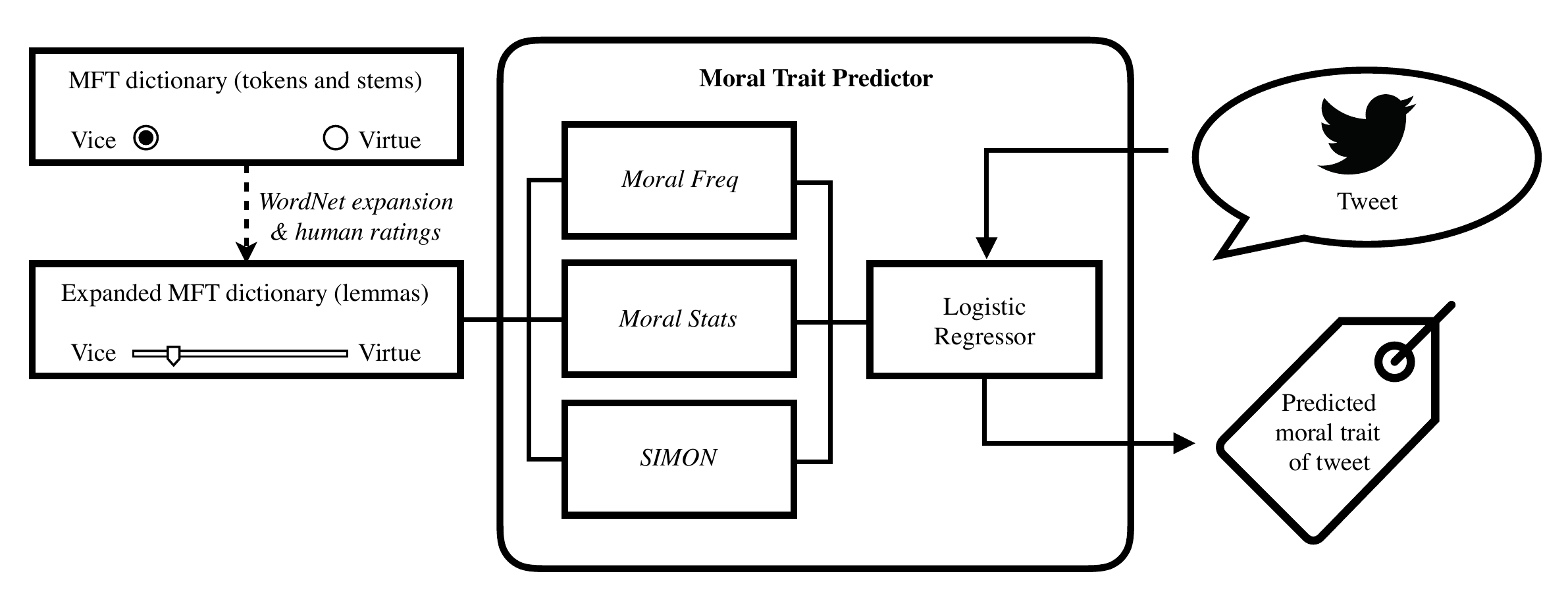}

\caption{Overview of the process, from dictionary expansion to moral value prediction}
\label{fig:overview}
\end{figure}
%\todo[color=blue!30]{Ideas or critiques for improving the graph? K: I LOVE the graphic :-) For improvement: 1) we show the expansion, but not how this defines the lexicons. 2) we have let's say 3 lego bricks being the 3 lexicons. they should appear somewhere between the mft dictionary boxes and the moral trait predictor. 3) the mft dictionary box no2 can be removed and on the arrow you can write both "expand with wordnet" and "human annotations". Oscar: In reference to the 3 boxes, the could appear inside the Moral Trait predictor. So, what about three arrows that come from the MFT dictionary and go to a classifier?}

\subsection{Expansion of The Moral Foundations Dictionary (MFD)}

The cornerstone of our study is the Moral Foundations Dictionary (MFD) \cite{Graham2009} which was created to capture the moral rhetoric according to the five predefined dimensions defined by the Moral Foundations Theory (MFT)  \cite{Graham2011}. 
The original MFD\footnote{Available at: \url{http://moralfoundations.org/othermaterials}} consists of lemmas and stems divided into ``virtue'' and ``vice'' \cite{Graham2009} for each foundation according to their moral polarity. ``Virtue'' words are foundation-supporting words (e.g., \textit{safe$^\star$} and \textit{shield} for Care ``virtue''), whereas ``vice'' words are foundation-violating words (e.g., \textit{kill} and \textit{ravage} for Care ``vice'').
MFD \cite{Graham2009} was meant to be used together with the Linguistic Inquiry and Word Count (LIWC) program \cite{Pennebaker2001}, and thus contains either lemmas (158 entries such as \textit{abandon}) or stems with a wild-card sign (166 listings), that LIWC analysis uses to match with all the forms of the base word; for instance, the entry \textit{abuse$^\star$} will match ``abuse'', ``abuses``, ``abused'', ``abuser'', ``abusers'', and so on. 
%The LIWC program was developed to count the proportion of words in different psychologically meaningful categories, and researchers have successfully applied it to a range of social psychology problems \cite{Pennebaker2001}.
%Graham et al.~\cite{Graham2009}, in their work, highlight the difficulty of creating a resource to assess morality due to a linguistic, cultural, and historical context which reflects on the language usage. 
Due to the limited amount of lemmas and stems of words, often radical or rarely used in everyday language, for instance, ``homologous'' and ``apostasy'', the expansion of the existing dictionary is of essential importance.

Since we are interested in lemmas instead of stems, we initially expanded the original dictionary using the  WordNet \cite{Miller1995} synsets, maintaining the lemmas that shared the same initial part with stems in the MFD. 
The result of this first expansion was to obtain for each MFD entry, for instance, \textit{traitor$^\star$}, a series of lemmas, for instance, \textit{traitor\#n}, \textit{traitorous\#a}, \textit{traitorously\#r}, \textit{traitorousness\#n}\footnote{The letter after the number sign \# indicates the part of speech for that word, i.e., \textit{\#n} for nouns, \textit{\#a} for adjectives, \textit{\#r} for adverbs, and \textit{\#v} for verbs.}.

We performed an initial preprocessing step on the obtained word corpus removing the forms that matched the search but did not relate to a moral trait.
For example, the stem \textit{caste$^\star$} not only matches \textit{caste\#n} and \textit{caste\_systems\#n}, but also \textit{caster\#n} and \textit{caster\_sugar\#n}, which are clearly not related to any moral foundation. This procedure was carried out manually, considering both the gloss for the lemma provided by WordNet and the moral trait that should be attributed to that word (e.g., while it could be argued that a statesman' name is an appropriate match for the Authority trait, the stem \textit{church$^\star$} relates to purity, and thus we ignored \textit{Churchill\#n}).

Following the original classification, we divided the obtained word corpus (1,148 words) in ``virtue'' and ``vice'' lemmas resulting with 520 ``virtues" and 476 ``vices" while 152 were characterized as ``general" morality words. 
These words can pertain to more than one traits; however, this is not common as the dataset consists of 442 unique ``vice" words and 512 unique ``virtue" words as shown in Table \ref{table:datasetsize}.

\begin{table}
\renewcommand{\arraystretch}{1.5}
\setlength{\tabcolsep}{12.2pt}
\centering
\begin{tabular}{lrr}
\toprule
Moral Dimension & Virtue  & Vice \\% & Total\\
\midrule
Care/Harm & 95 (16) & 85 (35) \\%& 180 (51)\\
Fairness/Cheating & 69 (26) & 57 (18) \\%& 126 (44)\\
Loyalty/Betrayal & 99 (29) & 72 (23) \\%& 171 (52)\\
Authority/Subversion & 160 (45)& 101 (37) \\%& 261 (82)\\
Purity/Degradation & 97 (35) & 161 (55) \\%& 258 (90)\\
\midrule
Total & 520 (151) & 476 (168)\\%& 996\\
\bottomrule
\end{tabular}
\caption{Corpus size after employing the WordNet resource to expand the MFD according to the official ``virtue'' and ``vice'' categories. The initial number of words contained in the MFD is shown in parenthesis.}
\label{table:datasetsize}
\end{table}

\subsection{Moral Valence Annotation}
\label{subsec:annotation}

Once the expanded dictionary was obtained, we used the Figure Eight\footnote{The Figure Eight Platform is available here: \url{https://www.figure-eight.com/}} crowdsourcing platform to annotate each lemma with an association strength to the related moral trait. The goal here is twofold. On the one hand, we can use these annotations to determine if the terms extracted during the expansion process are still related to a moral trait. On the other hand, a lexicon with ratings could be useful for better dictionary-based approaches and is a first step in the direction of moral detectors that can rank sentences, instead of merely classifying them with a binary vice/virtue rating.

The expanded dictionary was annotated in terms of moral valence, but we also collected ratings of valence and arousal, following the definitions employed for the ANEW resource~\cite{Miller1995}. For our purpose, moral valence can be represented by a bipolar scale that, in aggregate, defines a continuous dimension from one moral extremity of the MFT to the other, e.g., from Care to Harm. 
Moral valence was operationalized in a 9-point Likert Scale, wherein if a word was ranked in the middle of the scale, it was semantically neutral to the specific moral dimension.
The annotators were presented with the description of the moral trait and were asked to rate the relevance of the word to the specific foundation; if relevant, they were asked to rate its emotional valence, arousal and then its moral valence.
%Appendix \todo{O: remove this?} ~\ref{sec:app1} provides an overview of the rating procedure as an example of how the study was conducted. 
Each experiment presented 20 different words to the annotator.
The first time the annotators participated in the rating of a specific moral dimension (e.g., Care/Harm), after the experiment, they were asked to fill in the Moral Foundations Questionnaire \cite{Graham2009} for the respective dimension. At least five annotators were recruited for each lemma.

The ratings of valence and arousal were included to ensure a minimum quality of the annotation. Since no existing resource annotates moral valence on a fine scale, we used the values of valence from the subset of words that appear both in our extended dictionary and in \cite{Warriner}. Annotators have always been presented 4 ``gold'' words among the 20 words they annotate, and the annotations of those who fail more than 1 gold word are discarded. A valid answer is one that lies within 1.5 standard deviations from the valence mean of \cite{Warriner}, for each specific gold word.

As seen, the proposed lexicon has both subjective and generative components.
We need to take into consideration the subjectivity of human annotators; still, the candidate words for the annotations were chosen automatically, as a result of the expansion from a large word seed.

%Since this is the first resource of its kind, along with the lemmas of the expanded moral dictionary, we included in our annotation also some of the words of the ANEW database collected by Bradley and Lang \cite{Bradley1999}, to assess the quality and consistency of our annotations. \todo{Explain here that these words are used as gold standard, and annotators that end up being for two times were removed from the experiment and their annotations ignored.}

\subsection{Moral Lexicon Approaches}
\label{subsec:lexicon-use}

\begin{figure}[!ht]
  \centering
    \includegraphics[width=\linewidth]{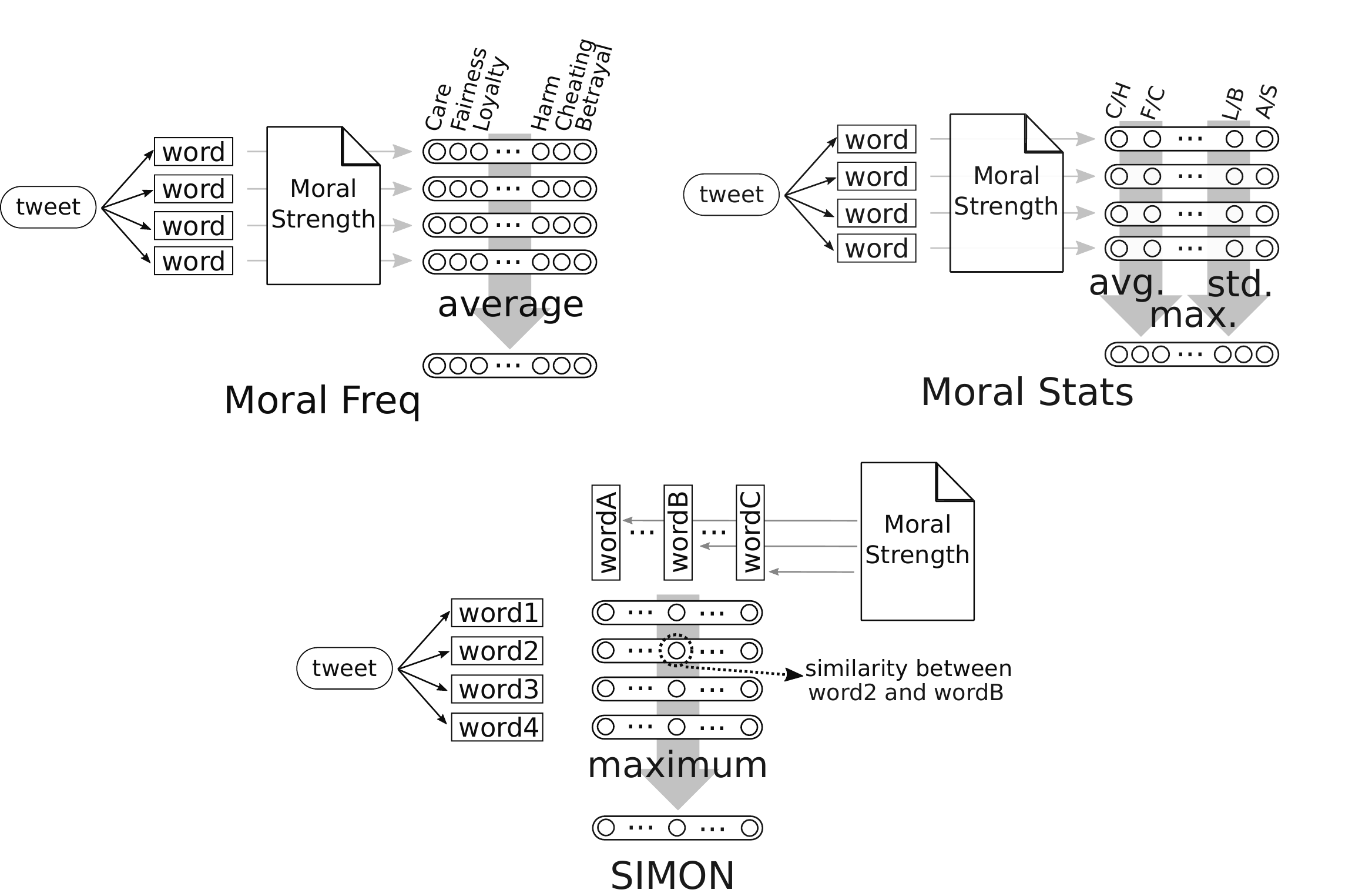}

\caption{Diagram of the proposed feature extraction approaches that utilize the presented lexicon.}
\label{fig:feature-extraction}
\end{figure}
For the generated moral lexicon, we propose the following feature extraction approaches, which can be divided into those that solely exploit the semantic information of each word, and those who exploit the moral valence associated to the word.
More specifically, we propose three lexicon utilization approaches:
(i) frequency counts,
(ii) statistical summary, and
(iii) word embedding similarity based representations.
The two first approaches use both the words and their moral values, while the third one makes use solely of the selection of words, ignoring the associated numeric moral values.
Figure~\ref{fig:feature-extraction} illustrates these approaches.
These three methods described above are used as feature extractors.
In the next experiments, we feed these features to a logistic regression classifier.
Such a simple learning algorithm is used to evaluate the performance of the proposed features, without exploiting more complex learning methods.

\textbf{\textit{Moral Freq}.} It consists of counting the number of words that express a specific moral dimension in a binary way.
To decide if a specific word expresses a moral, we apply a simple rule:
if the word has its moral value lower than a certain threshold, it does not convey that moral; if higher, the word does express that moral.
Given the properties of the generated moral lexicon, the threshold is set at 5.
We represent a given text with a 10-dimensional vector, which contains the corresponding normalized frequency counts, each for each moral extremity; for instance, \textit{care/harm} are represented by two dimensions, one for \textit{care} and other for \textit{harm}.

\textbf{\textit{Moral Stats}.} Given a specific text, we generate a statistical summary of the moral valence distribution of the contained words in the text.
In the statistical summary, we included (i) the average,
(ii) the standard deviation, (iii) the median, and (iv) the maximum value.
As a result, the text is represented by a 20-dimensional vector consisting of the statistical values obtained from the lexicon annotations.

\textbf{\acf{simon}.} Finally, the third approach is known as \acf{simon}, described in \cite{araque2018similarity}.
This method was initially developed for sentiment analysis tasks, while here, we adapted it to moral valence assessment.
\textit{SIMON} uses a pre-trained word embedding model to compute the cosine similarity between the words of the analyzed text and a selection of domain-related words, in our case, a specific moral dimension. 
Projecting the analyzed text over the selection of words from \textit{MoralStrength}, we result with a vector representation that encodes the similarity of the document to the specific moral dimension.
%That is, for each moral trait, the words extracted from the proposed moral lexicon have been used as the set of selected words for the \acs{simon} method. \todo{K: what do we mean here?} \todo{Oscar: it is actually a repetition of what has been said. I have tried to improve it}\todo{K: ok, got it.}

\subsection{Data Collection and Preprocessing}
\label{subsec:twitter-data}

For the evaluation of our models, we employed the Moral Foundation Twitter Corpus (MFTC) \cite{Hoover2019}.  MFTC is the most extensive available corpus containing 35,108 tweets and annotations, specifically collected to assess the moral values from user-generated content.
This corpus includes seven distinct datasets\footnote{The full dataset is available at \url{https://osf.io/k5n7y/}.} which were employed in scientific studies to assess the moral narratives in the user-generated text according to the moral foundations' theory. 
Here we provide an overview of the datasets that are included in the MFTC and are employed in our analysis.

\textbf{Hurricane Sandy (HS).} The first dataset we employed is presented in~\cite{Lin2018} and originally consisted of  4,191 tweets\footnote{This dataset can be obtained from \url{https://osf.io/nzx3q/}.}.
These Tweets contain hashtags relevant to the ``Hurricane Sandy'', a hurricane that caused significant damage to the Eastern seaboard of the United States in 2012. Due to Twitter regulations, the original dataset could not be fully recovered, leaving us with only 3,853 messages. We further removed the retweets, keeping only the original messages, to avoid overfitting the data.
In this way, the processed dataset consists of 3,478 instances.

\textbf{Baltimore Protest (BP).}
% automatic annotations
The second dataset is comprised of %18 million tweets~\cite{Mooijman2018}.
messages related to the 2015 Baltimore Protests, which were motivated by the death of Freddie Gray.
An older version of this dataset exists which contains a more significant number of instances.
Nevertheless, since the annotations of this older version are obtained by automated means~\cite{Mooijman2018}, we have decided to use the newer version, which has manual annotations.
%These data have been automatically annotated by a machine learning classifier, as described in~\cite{Mooijman2018}, and is freely available for research purposes.
%From this dataset, we randomly selected a subset of it, retaining 73,060 messages\footnote{We report only the statistics of the subset used in this study.}.
%We opted for such selection to reduce computational complexity\footnote{Namely, when performed over the whole 18 million instances, the experimental design (described in section~\ref{sec:evaluation_results}) would take over 100 days for the entire computation and testing.
%For the restricted dataset, the complete evaluation takes no longer than 10 hours.}.

\textbf{All Lives Matter (ALM).} Include \#BlueLivesMatter and \#AllLivesMatter hashtags and were posted between 2015-2016. These tweets were purchased from a third-party vendor.

\textbf{Black Lives Matter (BLM).} Posted between 2015-2016 about the Black Lives Matter Movement. Hashtags used to compile the corpus: \#BLM, \#BlackLivesMatter. The tweets were purchased from a third-party vendor.

\textbf{2016 Presidential Election (PE).} Scraped during the 2016 Presidential election season from the followers of $@$HillaryClinton, $@$realDonaldTrump, $@$NYTimes, $@$washingtonpost, and $@$WSJ.

\textbf{Davidson (D).} Taken from Davidson et al.'s~\cite{Davidson2017} corpus of hate speech and offensive language\footnote{The original corpus is available at \url{https://github.com/t-davidson/hate-speech-and-offensive-language/tree/master/data}.}.

All the above datasets are annotated by experts who indicated the presence or absence of a moral foundation dimension for each tweet.
Moreover, annotations include a ``non-moral'' label, indicating that the specific text does not reflect any moral trait.
% K: NOTE HERE
%{\color{red}A shortcoming of the \textit{Baltimore Protest} compared to the rest of datasets that is extremely important to consider is that the annotations included are not the result of human experts, but rather the output of a machine learning system.
%Thus, the annotations' quality is bounded by the error associated with the method proposed in the respective study~\cite{Mooijman2018}.
%In light of this, the \textit{Hurricane Sandy} dataset remains the point of reference since it does not entail intrinsic errors of any automatic system, but rather the opinion of human annotators.}
Importantly, the above datasets cover a wide variety of topics, both political and not. 
Topics related to politics cover left (e.g. BLM), right(e.g. ALM), and bilateral sides (e.g. Presidential Elections); while the datasets that are not unrelated to politics are expressing two controversial situations, a humanitarian call (e.g., Hurricane Sandy) and a collection of hate speech (e.g. Davidson). 
Such variability in the topics allows for a broader evaluation of the models' performance, avoiding biases due to context-specific language usage.

All data were collected downloading the original tweets following the Twitter IDs provided in the MFTC~\cite{Hoover2019}.
Since users often delete their tweets, we only managed to recover a portion of the original datasets. 
More specifically,  82\% of the original dataset has been recovered, and the statistics are reported in Table~\ref{tab:twitter-datasets}. We also report the distribution of Tweets per moral dimension per dataset.
We applied some basic preprocessing to the original textual content of the tweets employing
 the GSITK library\footnote{\url{https://github.com/gsi-upm/gsitk}}.
In particular, we normalized the text converting the URLs using the special token ``$<$url$>$'', usernames to ``$<$username$>$'', and hashtags to the token $<$hashtag$>$ and the word that is included in the hashtag (e.g., ``\#Baltimore'' to ``$<$hashtag$>$ Baltimore'').
Moreover, punctuation, symbols, and numbering were normalized.

\begin{table}[h]
\renewcommand{\arraystretch}{1.5}
\setlength{\tabcolsep}{3pt}
\centering
\begin{tabular}{l|rrrrrr}
\toprule
\textbf{Moral} &  \textbf{Sandy} & \textbf{Baltimore}  & \textbf{ALM} & \textbf{BLM} & \textbf{Elections} & \textbf{Davidson}\\
\midrule
Care & 217 & 434 & 1,314 & 1,065 & 798 & 462 \\
Fairness & 416 & 292 & 723 & 940 & 736 & 133 \\
Loyalty & 410 & 895 & 408 & 531 & 286 & 331 \\
Authority & 155 & 120 & 274 & 494 & 177 & 1,064 \\
Purity & 38 & 37 & 182 & 254 & 349 & 122 \\
No moral & 2,242 & 2,396 & 585 & 1,056 & 2,020 & 2,846 \\
\hline
Total & 3,478 & 4,174 & 3,486 & 4,340 & 4,366 & 4,958 \\
\hline
Under-sampling & 824 & 1,185 & 1,162 & 1,146 & 1,455 & 1,400 \\
\bottomrule
\end{tabular}
\caption{Statistics of the Moral Foundations Twitter Corpus employed in this study as a benchmark. All datasets were annotated by human annotators. %Therefore, even if smaller in size, \textit{Hurricane Sandy} is considered as the dataset of reference for this study.
Moreover, we note that according to the topic the distribution of the traits is varies.
Last row reports the average number of training instances when using under-sampling.
}
\label{tab:twitter-datasets}
\end{table}

\subsection{Experimental Design}
\label{sec:experimental_design}
To evaluate the potentials of the \textit{MoralStrength} lexicon, we postulate the problem as a classification task.
In particular, we employ the three approaches previously described, namely \textit{Moral Freq}, \textit{Moral Stats}, and \textit{SIMON}, to predict the moral rhetoric in each of the aforementioned datasets (see section~\ref{subsec:twitter-data}).

In our experimental design, we include a basic Bag-of-Words (unigram) model providing a standardized way of obtaining a baseline in the computational linguistics field. We also report as a baseline the frequency counts employing the original MFD.
%In brief,  the unigram model provides a representation of the given text in terms of a first-order Markovian model, predicting the next word in the sequence. 
We built a series of logistic regression models; firstly, we assess the predictive power of the unigrams, \textit{Moral Freq}, \textit{Moral Stats}, and \textit{SIMON} lexicon methods alone. Then, we train logistic regression models concatenating the features extracted by the above approaches\footnote{For replicability purposes, we have liberated the \textit{MoralStrength} lexicon along with the implementation of the presented methods in a GitHub repository\footnote{\url{https://github.com/oaraque/moral-foundations}}.}. In this way, we examine the effective performance of both engineered and word embedding features in analyzing user-generated text. We also combine the unigrams to the proposed lexicon approaches described above.
Hence, for each dataset and moral dimension, we train a series of logistic regression models following a 10-fold cross-validation scheme.
We then report the F1-score as the evaluation metric per moral dimension since this is the once employed in the majority of the related studies.

To directly compare our proposed framework with the current state-of-the-art approach of Lin et al.~\cite{Lin2018}, we replicated their same configuration.
Namely, we perform over-sampling on the original dataset to overcome the highly imbalanced nature of the benchmark data (see Section~\ref{subsec:twitter-data}).
After over-sampling on the \textit{Hurricane Sandy} data, we resulted with an average number of training examples, $N=6,128$, instead of the original dataset size, $N=3,478$ (see Table~\ref{tab:twitter-datasets}).

Since over-sampling implies ``artificial" data samples, we propose an alternative methodology; more specifically, we performed under-sampling, which also deals with the issue of unbalanced classes, however, in doing so, it randomly excludes data points of the most populated class. 
In this way, for the \textit{Hurricane Sandy} we had $N=824$ data points (see Table~\ref{tab:twitter-datasets}.
%, while for the \textit{Baltimore Protest} only $N=24,026$ out of the original $N=73,060$ remained.
By reporting the score for both methods, we ensure the results are not biased by the technique used to address the class imbalance.

For all experiments, we report the performance in terms of F1-score, which is the metric also employed by Lin et al.~\cite{Lin2018}, as well as the average F1-score over all moral dimensions.
Moreover, to compare the improvement of the simplest model, which for this study we consider being the \textit{Moral Freq} model, 
we employ the Friedman statistical test~\cite{demvsar2006statistical}, which yields a ranking of the proposed method ordered by their performance.
To obtain further insights on the statistical significance of our obtained results for the baseline model the Bonferroni-Dunn~\cite{demvsar2006statistical} post-hoc statistical test is performed with $\alpha=0.05$.

%Additionally, already trained models are also available to be used directly on textual data.

\section{Results and Discussion}
\label{sec:evaluation_results}

\subsection{Evaluation of Moral Valence}
\label{subsec:annotation_evaluation}
% \begin{itemize}
% \item Once the collection was done, we wanted to evaluate the quality of the collected ratings with an intrinsic evaluation
% \item however, the only dictionary available for MFT has only binary vice/virtue ratings
% \item so we binarised everything (1-9 Likert scale, $<4.5$ is a vice and $>5.5$ is a virtue, the rest are "neutral-ish")
% \item and compared the words in the original MFT dictionary with our binarised annotations
% \item used Cohen's Kappa for the comparison
% \item good results (Table \ref{tab:cohen}), all very much above the chance level.
% \end{itemize}

After collecting the moral valence ratings, we assessed the quality of the crowdsourced data with an intrinsic evaluation. However, since the only dictionary currently available for MFT has binary annotations (vice/virtue), a direct comparison with it is not informative enough.

Hence, we evaluate the quality by (i) calculating inter-annotator agreement for the moral valence ratings, (ii) calculating the correlation between valence scores and the normative lexicon of Warriner et al.~\cite{Warriner}, and (iii) comparing binarized moral valence ratings with the gold standard given by the MFD. The results for all these tests are reported in Table \ref{tab:agreement}.

To assess inter-annotator agreement we calculated Gwet's agreement coefficient (AC2) \cite{gwet2014handbook}. We opted for this measure since other, more common measures (e.g., Cohen's Kappa) require the number of annotators per element to be constant, and this is not the case for our data. Moreover, Gwet's coefficient can be weighted, meaning that annotators expressing close ratings will positively influence the coefficient score, and negatively impacted by scores that are far apart, a sensible feature for our dataset.
Results for all the traits are in the ``Moderate'' to ``Good'' range (0.4-0.8), except for fairness (which had ``Poor" agreement, 0.17). While this is positive, it also indicates that the task is not trivial and that some words might be hard to rate.
The lower agreement of fairness led us to inspect the agreements for all traits manually, and we discovered that some annotators were particularly inaccurate. It was thus decided to discard some annotators, despite their ability to complete the crowdsourced experiment without failing the control questions. In particular, for the Authority trait, the annotator with the worst agreement was removed, improving the original AC2 of 0.41 to 0.42. For loyalty, the answer of one annotator was lost due to programmatic error (the result for one word is outside the range specified by the Likert scale) and was removed from the dataset (no effect on the agreement). In the case of fairness, we intervened more drastically and removed five annotators, plus one non-valid answer. The five discarded annotators were chosen due to them having a weak agreement with other annotators, and to inconsistent ratings (i.e., they gave the same score to antonyms that have opposite traits in the MFD gold standard, such as ``honest'' and ``dishonest'').
The inter-annotator agreement for valence ranges between 0.61 and 0.72, thus falling in the ``Good'' category for the set of words of every moral trait. %, while that for arousal is between 0.52 and 0.7, except for the words from Purity (0.36). 
This indicates, in general, that annotating valence is easier and less controversial than rating moral traits.

We also compared the aggregated values of valence ratings (i.e., the mean of all valence annotations for a word) with the gold scores provided by \cite{Warriner}. In this case, we report the results of the Pearson correlation, which ranges from 0.79 to 0.95, indicating once again that the crowdsourced annotation is of good quality, and that differences between annotators are  within the acceptable range.

Finally, to be able to compare with the only gold standard for moral foundations, i.e., the Moral Foundations Dictionary, we binarized the aggregated annotations and excluded those whose average is 5 (the center of the Likert scale, meaning that the word is neither positive nor negative\footnote{While it would be sensible to consider neutral a range instead of a single value, e.g., excluding everything in the interval 4.5-5.5, we wanted to avoid removing more words from the comparison.}). This way, we could calculate Cohen's kappa coefficient by comparing to the vice/virtue ratings of the MFD for the subset of words that exists in both datasets. The lowest agreement is for Authority, but also, in this case, the 0.78 value suggests that the annotations are generally reliable and entirely in line with the original MFD. It is perhaps worth noting that the agreement of fairness is quite good (0.84), despite the lower inter-annotator agreement of the collected ratings. This might indicate that, while the aggregate ratings are reliable (i.e., they fall in the correct side of the morality spectrum), there is a relatively high individual variation regarding where the words of that dimension should be placed.

% Since MFD only includes binary labels (i.e., a word can either be a ``virtue'' or a ``vice''), to compare against it, we binarised the obtained scores. 
% For each moral trait, all words rated 4.5 or lower were assigned the \textit{vice} label, while all words higher than 5.5 were considered a \textit{virtue}. Words rated between 4.5 and 5.5 were removed from the comparison, as the exact threshold between \textit{vice} and \textit{virtue} is expected to revolve around 5, since we are on a 9-point Likert scale, but not necessarily 5.
% This binarisation allowed us to calculate the agreement between annotators' and the MFD ratings. Table \ref{tab:agreement} reports the agreement scores in terms of Cohen's Kappa, showing that the annotations are in line with the MFD standard.  \textit{Fairness}, the trait with the lowest agreement, is still considerably above the chance level which allows us to conclude that the obtained annotations are of good quality and can be employed as a ground-truth reference.

\begin{table}
\renewcommand{\arraystretch}{1.5}
\setlength{\tabcolsep}{11.2pt}
\centering
\begin{tabular}{lrrr}
\toprule
Moral trait & Inter-annotator & Warr correlation & MFD agreement\\
\midrule
Authority    &    0.42 & 0.84 & 0.78 \\
Care         &    0.65 & 0.95 & 0.91 \\
Fairness     &    0.34 & 0.88 & 0.84\\
Loyalty      &    0.59 & 0.91 & 0.84\\
Purity       &    0.56 & 0.79 & 0.92\\
\bottomrule
\end{tabular}
\caption{Measures of the quality of the collected ratings. The first column is the inter-annotator agreement for each moral dimension via Gwet's gamma with quadratic weighting metric. The second column is the correlation of the aggregate valence ratings and the gold standard of \cite{Warriner}. The last column is the agreement of the aggregate ratings (binarized) and the original Moral Foundations Dictionary, using Cohen's Kappa.}
\label{tab:agreement}
\end{table}

% \begin{table}
% \renewcommand{\arraystretch}{1.5}
% \setlength{\tabcolsep}{11.2pt}
% \centering
% \begin{tabular}{lr}
% \toprule
% Moral Trait & Cohen's Kappa \\
% \midrule
% Care     &    0.79    \\
% Fairness &    0.05\\ % need to remove CFIDs 18666417, 38202325 and then it goes up to .37
% Loyalty     &    0.62 \\
% Authority&    0.63 \\
% Purity      &    0.46 \\ % need to remove CFIDs 30695793, and it goes up to .61
% \bottomrule
% \end{tabular}
% \caption{Inter-annotator agreement (Gwet's gamma with quadratic weighting)
% \todo{\color{red}Lorenzo please merge the tables 3 and 4.}}
% \label{tab:gwen}
% \end{table}

% \subsection{evaluation of the embeddings - cross validation}
% We trained one classifier per trait following a {\color{red} i don't remember how many folds we have}-fold cross-validation technique.
% The obtained accuracies per trait are reported in the Table~\ref{tab:acc}.

% \subsection{evaluation of the embeddings - annotation 2}
% We opted for an additional evaluation of the classification models. We randomly selected {\color{red} say how many} lemmas from the expanded corpus for which we have no initial annotations, and we repeated the same annotation scheme.
% Then we compared in terms of mean square error {\color{red} or correlation?} the distance of the obtained moral valence as emerged from the classifier with respect to the opinion of the annotators. 

\subsection{Evaluation of \textit{MoralStrength} Lexicon}
\label{subsec:lexicon-use-evaluation}

In this section, we assess the predictive power of the various approaches exploiting \textit{MoralStrength} to analyze the moral rhetoric on the benchmark datasets described above.
We confront the performance of the models against a series of baseline models.
Initially, for each dataset, we report the performance of the model employing as features frequency counts from the MFD; this model shows how well the MFD alone would perform.
%Such simple approach using the original moral lexicon is used as comparison against all the other introduced methods.
Then, we report the performance of the predictive model employing unigrams, which provide an assessment of the difficulty of the task itself and the State-Of-The-Art (SOTA) performance for each dataset when available.

We show the performance of the logistic regression models of increasing complexity, starting from the \textit{Moral Freq, Moral Stats}, and \textit{SIMON}, followed by aggregations of the above lexicons.
For all experiments, we report the Friedman statistical test~\cite{demvsar2006statistical}, which yields a ranking of the proposed methods ordered by their performance.
To obtain further insights on the statistical significance of our obtained results to the baseline model the Bonferroni-Dunn~\cite{demvsar2006statistical} post-hoc statistical test is performed.
Note that in this study, for the statistical significance test, we employed the \textit{Moral Freq} with MFD model as a baseline model, and not the unigram one, since it is the one that infers on the simplest generated lexicon.

For the case of Hurricane Sandy we can see that across all moral dimensions, the model inferring on the aggregated unigram and \textit{\acs{simon}} features emerges as the best performing approach;  with a statistically significant improvement of the average F1-score - 87.6 over 62.4 reported by Lin et al.~\cite{Lin2018} (see 
Table~\ref{tab:lexicon-use-over-sampling}).
In this case alone, we employed over-sampling to directly compare to the previous SOTA approach on the same dataset~\cite{Lin2018}.
Interestingly, the highest score is obtained for ``purity'', which was reported being the most challenging moral dimension in the work of Lin et al.~\cite{Lin2018}.
Examining each moral dimension separately, we note that our results are also consistently higher than the unigram model.
The models that stand out are (i) unigrams + \textit{SIMON} for fairness, loyalty, and purity, (ii) unigrams + \textit{SIMON} + \textit{Moral Freq} for care, purity, and neutral text, while (iii)  unigrams + \textit{SIMON} + \textit{Moral Freq} + \textit{Moral Stats} is the best performing models for authority and purity. 

Table~\ref{tab:lexicon-use-under-sampling-1} reports the results of the evaluation for Hurricane Sandy when under-sampling is applied.
Following this sampling approach, the results vary for oversampling (see Table~\ref{tab:lexicon-use-over-sampling}), while the average overall performance 
slightly decreases (85.0\% against 87.6\% F1-score).
Still,  it is arguably preferable to perform under-sampling in comparison to over-sampling since in this way we avoid overfitting to the most prevalent outcome.
Noteworthy is the fact that the best performing models are consistent between the over and under-sampling approaches.
We also note that the importance of the statistical features regarding the moral valence of lemmas, exploited in the \textit{Moral Freq} and \textit{Moral Stats} lexicon methods, is more pronounced for all moral traits, for the oversampling technique.
More precisely, the model inferring from unigrams together with the \textit{Moral Stats} model, has a better performance in fairness, loyalty, and purity, while for care, the best performing model is the \textit{\acs{simon}} combined to the \textit{Moral Freq} and \textit{Moral Stats} Lexicons. 
Observing the obtained results, we conclude that combining lexicon-driven representations which take into consideration the moral valence, together with pure textual information (for instance, the unigrams), allows for a more robust and semantically meaningful representation. 
%discuss the third paper - see where this fits best.
Despite the differences in the proposed approaches, comparing our approach to the study presented by Garten et al.~\cite{Garten2016}, who also predicted the moral foundations on the  \textit{Hurricane Sandy} dataset, we note that their best performing model achieved 49.6\% F1-Score, which is remarkably lower than the 88.2\% reported here.

% discuss the Baltimore Protest
%\textbf{Evaluation on Baltimore Protest.}
Next we present the performance of the models on  \textit{Baltimore Protest}, \textit{All Lives Matter}, \textit{Black Lives Matter}, \textit{Davidson} and \textit{2016 Presidential Election} datasets in Tables~\ref{tab:lexicon-use-over-sampling}, \ref{tab:lexicon-use-under-sampling-1}, \ref{tab:lexicon-use-under-sampling-2}, \ref{tab:alm}, \ref{tab:blm}, \ref{tab:davidson}, and \ref{tab:election}, respectively. 
According to the previous discussion, we employ the under-sampling technique in all cases, always reporting the SOTA, as described in Hoover et al.~\cite{Hoover2019}
\footnote{Regarding \textit{Baltimore Protest}, Rezapour et al~\cite{Rezapour2019} reported higher accuracy with respect to Mooijman et al.~\cite{Mooijman2018}.
Nevertheless, we cannot compare against them for two reasons.
First, the dataset is not the same;  Rezapour et al. selected a subset from the original larger dataset~\cite{Mooijman2018} where annotations were automatically inferred by an algorithm, while our evaluation dataset originates by Hoover et al~\cite{Hoover2019}, where annotations were manually assigned.
Secondly, their evaluation metric is reported in terms of accuracy, while we use F1-Score as the majority of the related works.
Thus, a direct comparison can not be made.}.
%So, this result has not been included in the statistical test, but still, we show it in Table~\ref{tab:lexicon-use-under-sampling-2} to facilitate a rough comparison to the reader.
%{\color{red} I would put Hoover in the table 6 in this case as in everything else. this one is f-score right?}
Carefully comparing the experimental results, we first note that unigrams provide a reasonably good baseline for all datasets.
As expected,  this method is shown to be a generally stable approach, even if the training samples are few.
Next, we note that when we introduce the notion of moral strength to the basic unigram approach, results are steadily improved.
Table~\ref{tab:all_datasets}, provides clear evidence to this statement; there, the Friedman test indicates that the best model overall is unigrams + \textit{Moral Freq}, followed by unigrams + \textit{Moral Stats}.
Thus, introducing knowledge about moral valence, we can better predict the moral rhetoric in text.

This result differs from the statement made by Lin et al.~\cite{Lin2018}, where they show that adding the features from the original lexicon, MFD, does not improve the score.
Hence, we could argue that introducing the notion of moral valence the quality of the proposed lexicon, \textit{MoralStrength}, improves the performance in text analysis as compared to the MFD.
To safely conclude to this latest argument, we present a direct comparison of all datasets in Table~\ref{tab:baselines_comparison}.
Here, we compare the performance of the original MFD versus our \textit{MoralStrength} for all the proposed approaches. The reported scores are averaged over all moral values per dataset.
As observed, the Friedman test indicates that the \textit{SIMON} model with the proposed lexicon outperforms the rest.
Hence, it is safe to assume that the newly introduced resource offers an improvement over the previous lexicon.

Moving to the model comparison, it can be seen that combining unigrams and \textit{SIMON} does not generally improve the results of the classification.
Interestingly enough, such combination does primarily improve the metrics when done in the over-sampling case (Table~\ref{tab:lexicon-use-over-sampling}).
In light of this contrast, and considering that both the unigrams and SIMON approach generate a large number of features, we hypothesize that combining large feature vectors leads to overfitting.
As expected, an increase of the training data quantity improves the results, leading the unigrams + SIMON model to obtain better results.

% O: I removed this, cause it was particular of sandy and baltimore
%We extended the evaluation, employing the \textit{Baltimore Protest}, which is presented in the study of Mooijman et al.~\cite{Mooijman2018}.
%Since their approach is not directly comparable to ours, we perform the validation process only on training data obtained with the under-sampling technique.
%Note that ground-truth here are annotations obtained via a neural network model, and hence, our model's evaluation is bounded to the performance of the system presented in Mooijman et al.~\cite{Mooijman2018}.
%We note that the overall performance for all moral dimensions is lower to the respective performances on \textit{Hurricane Sandy} (see Table~\ref{tab:lexicon-use-under-sampling-1}), with the sole exception the case of loyalty, with the average F1-score to be 87.2\%. 
%Similar as before, for all moral dimensions, the best performing model infers on the features obtained from the unigrams; in three out of five dimensions the best model is the one inferring on the combined with the \textit{SIMON} word embeddings, while for the remaining two, from the combination with \textit{Moral Freq} and \textit{Moral Stats} Lexicons.
%Of course, unigrams combined with the \acs{simon} representation are more effective with more massive datasets, since the resulting vector representations are larger and more generic; however, the importance of the information entailed in the moral valence still emerges.

To conclude, we observe that the performance trends are maintained; a unigram model is a robust approach, and adding information from \textit{MoralStrength} improves the prediction performance.
Noteworthy is the fact that the expression of moral sentiment can vary substantially according to the context. Variability in the model performances may also depend on the topic of discourse; the datasets employed for the evaluation include political left (e.g., BLM), right (ALM), both ideological poles (e.g., the Presidential election).
They also include topics unrelated to political discourses (e.g., Hurricane Sandy). 
Moreover,  the variability of training samples available for each trait may explain the differences in the model performance. 
%This heterogeneity makes out-of-domain prediction particularly difficult because expressions of moral sentiment in one domain will not necessarily generalize well to data drawn from a different domain. Accordingly, to help address this issue, we provide moral sentiment annotations for Tweets drawn from multiple, heterogeneous contexts.

%Additionally, the consistently high performance against both smaller and larger datasets further validates the quality of the proposed approaches.
%, establishing new state-of-the-art results in the task of moral trait estimation
%In light of the results, we observe that using unigrams combined with \acs{simon} features yields better results, except for the case of \textit{Hurricane Sandy} with under-sampling.
%These results suggest that when enough data is available, the combination of unigrams and \acs{simon} obtains the best performance.
%That is, combining simple text representations with the use of the vocabulary (without numerical annotations) of the \textit{Moral Lexicon}.
%Nevertheless, when the number of available data examples is small, the better approach is to combine unigrams with more straightforward lexicon-based representations that make use of the numeric annotations.
We believe that exploratory analysis will be useful for the ever-increasing studies on moral foundations since it presents a variety of approaches on how the moral lexicon we propose can be employed for the prediction of moral narratives from a text.

\begin{table*}
\centering
\addtolength{\leftskip}{-3cm}
\addtolength{\rightskip}{-3cm} %extend page margins to center table
\renewcommand{\arraystretch}{1.5}
\begin{tabular}{l|rrrrrr|r|r}
\toprule
\textbf{Approach} & \textbf{C/H} & \textbf{F/C} & \textbf{L/B} & \textbf{A/S} & \textbf{P/D} & \textbf{NM} & \textbf{Avg.} & \textbf{Rank} \\
\midrule
Baseline: Frequency MFD &    56.3 &     59.2 &    61.8 &      54.4 &    63.1 &      66.4 &    60.2 &    14.4 \\
unigrams &      74.0 &     76.9 &    76.5 &      80.7 &    94.1 &      77.2 &    79.9 &    11.9 \\
SOTA: Lin et al~\cite{Lin2018} &    82.3 &     70.7 &    50.3 &      69.3 &    37.4 &      64.2 &    62.4 &    12.9 \\
\hline
\textit{Moral Freq} &    61.4 &     58.2 &    61.9 &        56.0 &    62.1 &      63.4 &    60.5 &    14.4 \\
\textit{Moral Stats} &    62.8 &     57.2 &    58.8 &      52.7 &    64.1 &      63.3 &    59.8 &    15.1 \\
SIMON &    79.6 &     82.3 &    77.1 &        86.0 &    98.1 &      84.2 &    84.5 &    6.4$^\star$ \\
\hline
SIMON + \textit{Moral Freq} &    79.2 &     82.5 &    77.2 &      83.8 &    98.2 &      83.9 &    84.1 &    6.8$^\star$ \\
SIMON + \textit{Moral Stats} &    79.2 &     82.2 &      77.0 &        84.0 &    98.2 &      83.9 &    84.1 &     7.6 \\
SIMON + \textit{Moral Freq} + \textit{Moral Stats} &    79.6 &     82.5 &    77.1 &        84.0 &    98.2 &      83.8 &    84.2 &    6.6$^\star$ \\
\hline
unigrams + \textit{Moral Freq} &    75.3 &     77.7 &    77.2 &      81.2 &    95.5 &      77.8 &    80.8 &     9.7 \\
unigrams + \textit{Moral Stats} &    73.5 &     77.6 &    76.7 &      81.3 &    95.7 &      77.9 &    80.5 &    10.8 \\
unigrams + \textit{Moral Freq} + \textit{Moral Stats} &      74.0 &     78.2 &    77.1 &      81.7 &    95.9 &      77.9 &    80.8 &     9.4 \\
unigrams + SIMON &    84.6 &   \textbf{85.6} &  \textbf{81.2} &        90.0 &  \textbf{98.9} &      85.5 &  \textbf{87.6} &  \textbf{2.0}$^{\star\dagger}$ \\
unigrams + SIMON + \textit{Moral Freq} &  \textbf{85.1} &     85.2 &    80.8 &      89.5 &  \textbf{98.9} &    \textbf{85.6} &    87.5 &    2.6$^{\star\dagger}$ \\
unigrams + SIMON + \textit{Moral Stats} &    84.9 &     85.4 &    80.4 &        90.0 &    98.8 &      85.2 &    87.5 &    3.3$^{\star\dagger}$ \\
unigrams + SIMON + \textit{Moral Freq} + \textit{Moral Stats} &      85.0 &     85.4 &    80.8 &    \textbf{90.2} &  \textbf{98.9} &      85.2 &  \textbf{87.6} &    2.1$^{\star\dagger}$ \\
\bottomrule
\end{tabular}
\caption{F1-Score of the proposed methods using over-sampling over \textit{Hurricane Sandy} (\cite{Lin2018}). C/H: Care/Harm, F/C: Fairness/Cheating, L/B: Loyalty/Betrayal, A/S: Authority/Subversion, P/D: Purity/Degradation, NM: Non-moral, Avg.: Average. `$^\star$' and `$^{\dagger}$' mark that the approach significantly outperforms the MFD baseline and the SOTA, respectively. The model with the lowest rank is the one that outperforms the rest.}
\label{tab:lexicon-use-over-sampling}
\end{table*}

\begin{table*}
\centering
\addtolength{\leftskip}{-3cm}
\addtolength{\rightskip}{-3cm} %extend page margins to center table
\renewcommand{\arraystretch}{1.5}
\begin{tabular}{l|rrrrrr|r|r}
\toprule
\textbf{Approach} & \textbf{C/H} & \textbf{F/C} & \textbf{L/B} & \textbf{A/S} & \textbf{P/D} & \textbf{NM} & \textbf{Avg.} & \textbf{Rank}\\
\midrule
Baseline: Frequency MFD &    56.1 &     54.6 &    61.3 &      50.4 &    59.4 &      66.2 &      58.0 &    15.1 \\
unigrams &    78.1 &     88.8 &    85.7 &    \textbf{90.1} &    66.5 &    \textbf{92.1} &    83.5 &    5.3$^{\star\dagger}$ \\
SOTA:  Hoover et al~\cite{Hoover2019} &      55.0 &       58.0 &      44.0 &        44.0 &      56.0 &       - &    51.4 &    13.4 \\
\hline
\textit{Moral Freq} &   65.2 &     56.4 &    61.6 &      51.7 &    54.6 &      69.3 &    59.8 &    14.6 \\
\textit{Moral Stats} &    74.1 &     60.8 &    73.4 &      63.3 &    60.4 &      79.6 &    68.6 &    12.6 \\
SIMON &    76.9 &     76.3 &    77.2 &        75.0 &    73.5 &      72.8 &    75.3 &    10.7 \\
\hline
SIMON + \textit{Moral Freq} &    77.2 &     79.6 &    79.4 &      75.6 &    70.7 &      73.4 &      76.0 &    10.6 \\
SIMON + \textit{Moral Stats} &    77.3 &     80.8 &    80.2 &      74.6 &    72.2 &        76.0 &    76.8 &    10.1 \\
SIMON + \textit{Moral Freq} + \textit{Moral Stats} &    77.5 &     80.8 &    80.7 &      75.1 &    72.4 &      77.1 &    77.3 &     9.1 \\
\hline
unigrams + \textit{Moral Freq} &    81.2 &     88.4 &  \textbf{86.8} &      89.9 &    67.4 &      91.9 &    84.3 &    4.4$^{\star\dagger}$ \\
unigrams + \textit{Moral Stats} &    80.2 &     87.3 &    85.5 &        89.0 &    67.8 &      90.4 &    83.4 &    6.4$^\star$ \\
unigrams + \textit{Moral Freq} + \textit{Moral Stats} &      80.0 &     87.3 &    85.6 &      89.1 &    68.2 &      90.2 &    83.4 &    6.2$^{\star\dagger}$ \\
unigrams + SIMON &  \textbf{83.8} &   \textbf{92.0} &    85.2 &      89.1 &  \textbf{74.7} &        85.0 &  \textbf{85.0} &  \textbf{3.3}$^{\star\dagger}$ \\
unigrams + SIMON + \textit{Moral Freq} &    83.1 &     91.5 &      86.0 &      88.1 &    73.2 &      85.7 &    84.6 &    3.6$^{\star\dagger}$ \\
unigrams + SIMON + \textit{Moral Stats} &    82.5 &     90.6 &    84.8 &        86.0 &    72.8 &        88.0 &    84.1 &    5.4$^{\star\dagger}$ \\
unigrams + SIMON + \textit{Moral Freq} + \textit{Moral Stats} &    82.7 &     90.4 &    85.2 &      85.6 &    72.6 &      88.1 &    84.1 &    5.3$^{\star\dagger}$ \\
\bottomrule
\end{tabular}
\caption{F1-Score of the proposed methods using under-sampling over \textit{Hurricane Sandy} (\cite{Lin2018}). C/H: Care/Harm, F/C: Fairness/Cheating, L/B: Loyalty/Betrayal, A/S: Authority/Subversion, P/D: Purity/Degradation, NM: Non-moral, Avg.: Average. `$^\star$' marks that the approach statistically outperforms the baseline. The model with the lowest rank is the one that outperforms the rest.
}
\label{tab:lexicon-use-under-sampling-1}
\end{table*}

\begin{table*}
\centering
\addtolength{\leftskip}{-3cm}
\addtolength{\rightskip}{-3cm} %extend page margins to center table
\renewcommand{\arraystretch}{1.5}
\begin{tabular}{l|rrrrrr|r|r}
\toprule
\textbf{Approach} & \textbf{C/H} & \textbf{F/C} & \textbf{L/B} & \textbf{A/S} & \textbf{P/D} & \textbf{NM} & \textbf{Avg.} & \textbf{Rank}\\
\midrule
Baseline: Frequency MFD &    59.3 &     64.4 &    64.3 &      50.2 &    54.7 &      64.1 &    59.5 &    14.9 \\
unigrams &    87.3 &       84.0 &    85.6 &      83.7 &  \textbf{93.2} &      82.8 &    86.1 &    4.3$^{\star\dagger}$ \\
SOTA: Hoover et al~\cite{Hoover2019} &      33 &       47 &      25 &        25 &      15 &       - &      29 &    13.9 \\
\hline
\textit{Moral Freq} &    57.6 &     67.9 &    64.5 &        54.0 &    60.7 &      63.6 &    61.4 &      14.1 \\
\textit{Moral Stats} &    66.5 &     72.5 &      71.0 &      63.7 &    65.6 &      68.1 &    67.9 &    12.3 \\
SIMON &    79.3 &     73.6 &    74.5 &      87.1 &    75.5 &      81.9 &    78.7 &     9.9 \\
\hline
SIMON + \textit{Moral Freq} &      80.0 &     66.4 &    77.7 &    \textbf{87.5} &    86.3 &        82.0 &      80.0 &     9.4 \\
SIMON + \textit{Moral Stats} &    81.5 &     71.7 &    77.6 &      86.7 &    84.9 &      82.1 &    80.7 &     9.3 \\
SIMON + \textit{Moral Freq} + \textit{Moral Stats} &    81.5 &     73.8 &    77.6 &      85.8 &    84.9 &      82.1 &      81.0 &     8.7 \\
\hline
unigrams + \textit{Moral Freq} &    88.1 &     81.4 &    85.2 &      85.4 &    91.9 &        84.0 &      86.0 &    4.1$^{\star\dagger}$ \\
unigrams + \textit{Moral Stats} &      88.0 &   \textbf{85.0} &  \textbf{85.8} &      86.7 &    90.5 &      83.5 &    86.6 &    3.4$^{\star\dagger}$ \\
unigrams + \textit{Moral Freq} + \textit{Moral Stats} &  \textbf{88.5} &   \textbf{85.0} &    85.5 &      86.7 &    90.5 &      83.8 &  \textbf{86.7} &  \textbf{3.1}$^{\star\dagger}$ \\
unigrams + SIMON &      86.0 &     68.5 &    84.8 &      81.7 &    87.8 &      85.1 &    82.3 &    7.6$^\star$ \\
unigrams + SIMON + \textit{Moral Freq} &    86.4 &     65.4 &    84.2 &      81.2 &    90.5 &      85.1 &    82.1 &    8.2$^\star$ \\
unigrams + SIMON + \textit{Moral Stats} &    86.6 &     72.1 &    84.7 &      82.9 &    86.4 &    \textbf{85.7} &    83.1 &    6.8$^{\star\dagger}$ \\
unigrams + SIMON + \textit{Moral Freq} + \textit{Moral Stats} &    87.1 &     72.1 &    84.8 &      83.3 &    86.4 &    \textbf{85.7} &    83.3 &    6.1$^{\star\dagger}$ \\
\bottomrule
\end{tabular}
\caption{F1-Score of the proposed methods using under-sampling over \textit{Baltimore Protest} (\cite{Hoover2019}). C/H: Care/Harm, F/C: Fairness/Cheating, L/B: Loyalty/Betrayal, A/S: Authority/Subversion, P/D: Purity/Degradation, NM: Non-moral, Avg.: Average.  `$^\star$' and `$^{\dagger}$' mark that the approach significantly outperforms the MFD baseline and the SOTA, respectively. The model with the lowest rank is the one that outperforms the rest.
}
\label{tab:lexicon-use-under-sampling-2}
\end{table*}

% new tables
\begin{table*}
\centering
\addtolength{\leftskip}{-3cm}
\addtolength{\rightskip}{-3cm} %extend page margins to center table
\renewcommand{\arraystretch}{1.5}
\begin{tabular}{l|rrrrrr|r|r}
\toprule
\textbf{Approach} & \textbf{C/H} & \textbf{F/C} & \textbf{L/B} & \textbf{A/S} & \textbf{P/D} & \textbf{NM} & \textbf{Avg.} & \textbf{Rank}\\
\midrule
Baseline: Frequency MFD &    65.9 &     77.7 &    63.8 &      82.8 &    57.5 &        65 &    68.8 &    14.9 \\
unigrams &      74.0 &   \textbf{91.9} &    88.1 &      90.7 &      92.0 &    \textbf{79.7} &  \textbf{86.1} &  \textbf{3.6}$^{\star\dagger}$ \\
SOTA: Hoover et al~\cite{Hoover2019} &      67.0 &       76.0 &      62.0 &        63.0 &      39.0 &       - &    61.4 &    14.3 \\
\hline
\textit{Moral Freq} &    64.6 &     76.7 &    67.6 &      85.6 &    58.3 &      57.4 &    68.4 &    14.9 \\
\textit{Moral Stats} &    64.4 &     85.4 &    85.3 &    \textbf{93.8} &    70.6 &      60.3 &    76.6 &    11.6 \\
SIMON &      75.0 &     83.6 &    76.2 &      80.3 &    86.8 &      73.6 &    79.3 &    11.4 \\
\hline
SIMON + \textit{Moral Freq} &    72.4 &     86.1 &    78.1 &      78.6 &    87.9 &      71.7 &    79.2 &    12.4 \\
SIMON + \textit{Moral Stats} &    72.7 &     86.7 &      78.0 &      82.3 &    89.6 &      68.8 &    79.7 &    11.7 \\
SIMON + \textit{Moral Freq} + \textit{Moral Stats} &    72.3 &     87.7 &    79.9 &      81.7 &    90.1 &      69.2 &    80.2 &    11.1 \\
\hline
unigrams + SIMON + \textit{Moral Stats} &    74.9 &     90.1 &    84.9 &      85.2 &  \textbf{93.1} &      74.4 &    83.8 &    6.1$^\star$ \\
unigrams + \textit{Moral Freq} &    74.5 &     90.4 &    88.3 &      90.7 &    89.8 &      78.2 &    85.3 &    4.6$^{\star\dagger}$ \\
unigrams + \textit{Moral Stats} &    70.8 &     90.9 &    89.1 &      90.7 &    92.8 &      73.2 &    84.6 &    5.9$^{\star\dagger}$ \\
unigrams + \textit{Moral Freq} + \textit{Moral Stats} &    72.2 &     91.3 &  \textbf{89.3} &      90.7 &    92.8 &      73.6 &      85.0 &    4.9$^{\star\dagger}$ \\
unigrams + SIMON &  \textbf{75.6} &     88.5 &    86.1 &      87.2 &    90.6 &      73.3 &    83.6 &    6.5$^{\star\dagger}$ \\
unigrams + SIMON + \textit{Moral Freq} &    74.8 &     88.4 &    83.9 &      86.5 &    90.4 &      73.7 &    82.9 &     8.1$^\dagger$ \\
unigrams + SIMON + \textit{Moral Stats} &    74.9 &     90.1 &    84.9 &      85.2 &  \textbf{93.1} &      74.4 &    83.8 &    6.1$^{\star\dagger}$ \\
unigrams + SIMON + \textit{Moral Freq} + \textit{Moral Stats} &  \textbf{75.6} &     90.2 &    84.9 &      86.5 &  \textbf{93.1} &      74.2 &    84.1 &    5.0$^{\star\dagger}$ \\
\bottomrule
\end{tabular}
\caption{F1-Score of the proposed methods using under-sampling over \textit{ALM} \cite{Hoover2019}. C/H: Care/Harm, F/C: Fairness/Cheating, L/B: Loyalty/Betrayal, A/S: Authority/Subversion, P/D: Purity/Degradation, NM: Non-moral, Avg.: Average. `$^\star$' and `$^{\dagger}$' mark that the approach significantly outperforms the MFD baseline and the SOTA, respectively. The model with the lowest rank is the one that outperforms the rest.}
\label{tab:alm}
\end{table*}

\begin{table*}
\centering
\addtolength{\leftskip}{-3cm}
\addtolength{\rightskip}{-3cm} %extend page margins to center table
\renewcommand{\arraystretch}{1.5}
\begin{tabular}{l|rrrrrr|r|r}
\toprule
\textbf{Approach} & \textbf{C/H} & \textbf{F/C} & \textbf{L/B} & \textbf{A/S} & \textbf{P/D} & \textbf{NM} & \textbf{Avg.} & \textbf{Rank}\\
\midrule
Baseline: Frequency MFD &      68.0 &     84.3 &      89 &      89.3 &    82.1 &      70.6 &    80.5 &    15.4 \\
unigrams &    85.8 &     93.1 &    90.5 &      93.3 &    93.3 &    \textbf{81.6} &  \textbf{89.6} &  \textbf{3.0}$^{\star\dagger}$ \\
SOTA: Hoover et al~\cite{Hoover2019}  &      74.0 &       87.0 &      83.0 &        25.0 &      57.0 &       - &    65.2 &      14.0 \\
\hline
\textit{Moral Freq} &    66.2 &     86.2 &    88.8 &      92.3 &    80.8 &      70.9 &    80.9 &    13.9 \\
\textit{Moral Stats} &    77.3 &   \textbf{93.4} &  \textbf{92.8} &    \textbf{96.6} &    90.5 &      51.6 &    83.7 &    7.6$^\star$ \\
SIMON &    81.7 &     85.2 &    88.7 &      86.9 &    84.8 &      75.8 &    83.9 &    12.5 \\
\hline
SIMON + \textit{Moral Freq} &      81.0 &     88.9 &    89.6 &        89.0 &    85.4 &      71.7 &    84.3 &    12.2 \\
SIMON + \textit{Moral Stats} &    80.8 &     89.9 &    90.6 &      89.8 &    84.4 &      71.8 &    84.6 &    10.6 \\
SIMON + \textit{Moral Freq} + \textit{Moral Stats} &    80.9 &     89.8 &    90.6 &      90.1 &    87.2 &      72.5 &    85.2 &     9.4 \\
\hline
unigrams + SIMON + \textit{Moral Stats} &    83.2 &     90.1 &    90.5 &      90.9 &    87.2 &      75.1 &    86.2 &    7.9$^\star$ \\
unigrams + \textit{Moral Freq} &    86.1 &     91.6 &    89.6 &      93.1 &    92.3 &      79.9 &    88.8 &    4.6$^\star$ \\
unigrams + \textit{Moral Stats} &    79.3 &       92.0 &    89.8 &      92.2 &    93.7 &      73.3 &    86.7 &    7.4$^{\star\dagger}$ \\
unigrams + \textit{Moral Freq} + \textit{Moral Stats} &    80.3 &     92.3 &    89.8 &      92.2 &  \textbf{94.3} &      73.3 &      87.0 &    6.8$^\star$ \\
unigrams + SIMON &  \textbf{88.2} &     91.1 &    90.9 &      89.6 &    84.8 &      79.4 &    87.3 &    6.1$^{\star\dagger}$ \\
unigrams + SIMON + \textit{Moral Freq} &    87.7 &     88.7 &    90.5 &      90.1 &    90.3 &      76.4 &    87.3 &    6.7$^{\star\dagger}$ \\
unigrams + SIMON + \textit{Moral Stats} &    83.2 &     90.1 &    90.5 &      90.9 &    87.2 &      75.1 &    86.2 &    7.9$^\star$ \\
unigrams + SIMON + \textit{Moral Freq} + \textit{Moral Stats} &      84.0 &     91.4 &    90.5 &      90.7 &    87.6 &      75.5 &    86.6 &    6.9$^\star$ \\
\bottomrule
\end{tabular}
\caption{F1-Score of the proposed methods using under-sampling over \textit{BLM} \cite{Hoover2019}. C/H: Care/Harm, F/C: Fairness/Cheating, L/B: Loyalty/Betrayal, A/S: Authority/Subversion, P/D: Purity/Degradation, NM: Non-moral, Avg.: Average. `$^\star$' and `$^{\dagger}$' mark that the approach significantly outperforms the MFD baseline and the SOTA, respectively. The model with the lowest rank is the one that outperforms the rest.}
\label{tab:blm}
\end{table*}

\begin{table*}
\centering
\addtolength{\leftskip}{-3cm}
\addtolength{\rightskip}{-3cm} %extend page margins to center table
\renewcommand{\arraystretch}{1.5}
\begin{tabular}{l|rrrrrr|r|r}
\toprule
\textbf{Approach} & \textbf{C/H} & \textbf{F/C} & \textbf{L/B} & \textbf{A/S} & \textbf{P/D} & \textbf{NM} & \textbf{Avg.} & \textbf{Rank}\\
\midrule
Baseline: Frequency MFD &    36.5 &     33.3 &    39.4 &      36.1 &    38.7 &        38.0 &      37.0 &    14.8 \\
unigrams &    84.6 &   \textbf{91.3} &    86.6 &    \textbf{77.3} &    92.3 &    \textbf{56.5} &  \textbf{81.4} &  \textbf{4.4}$^{\star\dagger}$ \\
SOTA: Hoover et al~\cite{Hoover2019} &       7.0 &        5.0 &       2.0 &         2.0 &       5.0 &       - &     4.2 &    13.9 \\
\hline
\textit{Moral Freq} &    39.6 &     33.3 &    39.9 &      39.1 &    37.9 &      39.6 &    38.2 &    14.3 \\
\textit{Moral Stats} &    42.4 &       39.0 &      43.0 &      43.1 &    37.9 &      42.7 &    41.3 &    13.4 \\
SIMON &    84.8 &     85.8 &    87.2 &        72.0 &    89.3 &      53.8 &    78.8 &    11.2 \\
\hline
SIMON + \textit{Moral Freq} &    85.4 &     85.8 &    87.7 &      71.1 &    89.8 &      54.3 &      79.0 &     9.4 \\
SIMON + \textit{Moral Stats} &    85.7 &     85.8 &  \textbf{88.4} &      72.4 &    89.8 &      54.5 &    79.4 &    7.6$^\star$ \\
SIMON + \textit{Moral Freq} + \textit{Moral Stats} &    85.7 &     85.8 &  \textbf{88.4} &      72.4 &    89.8 &      54.4 &    79.4 &     7.8 \\
\hline
unigrams + \textit{Moral Freq} &    84.6 &     90.6 &    87.6 &      76.2 &    92.3 &        56.0 &    81.2 &    4.9$^{\star\dagger}$ \\
unigrams + \textit{Moral Stats} &    85.4 &   \textbf{91.3} &    87.4 &        77.0 &    90.6 &      55.8 &    81.2 &    4.9$^{\star\dagger}$ \\
unigrams + \textit{Moral Freq} + \textit{Moral Stats} &    85.4 &   \textbf{91.3} &    87.4 &        77.0 &    90.6 &      55.7 &    81.2 &    5.1$^{\star\dagger}$ \\
unigrams + SIMON &    83.7 &     86.5 &      88.0 &      74.8 &    91.5 &      54.2 &    79.8 &     8.1 \\
unigrams + SIMON + \textit{Moral Freq} &    85.3 &     86.5 &      88.0 &      75.2 &  \textbf{92.8} &      54.9 &    80.5 &    6.0$^{\star\dagger}$ \\
unigrams + SIMON + \textit{Moral Stats} &    86.3 &     87.2 &    87.6 &      75.4 &    91.9 &      55.2 &    80.6 &    5.2$^{\star\dagger}$ \\
unigrams + SIMON + \textit{Moral Freq} + \textit{Moral Stats} &  \textbf{86.4} &     87.2 &    87.6 &      75.5 &    91.9 &        55.0 &    80.6 &    5.1$^{\star\dagger}$ \\
\bottomrule
\end{tabular}
\caption{F1-Score of the proposed methods using under-sampling over \textit{Davidson} \cite{Hoover2019}. C/H: Care/Harm, F/C: Fairness/Cheating, L/B: Loyalty/Betrayal, A/S: Authority/Subversion, P/D: Purity/Degradation, NM: Non-moral, Avg.: Average. `$^\star$' and `$^{\dagger}$' mark that the approach significantly outperforms the MFD baseline and the SOTA, respectively.  baseline. The model with the lowest rank is the one that outperforms the rest.}
\label{tab:davidson}
\end{table*}

\begin{table*}
\centering
\addtolength{\leftskip}{-3cm}
\addtolength{\rightskip}{-3cm} %extend page margins to center table
\renewcommand{\arraystretch}{1.5}
\begin{tabular}{l|rrrrrr|r|r}
\toprule
\textbf{Approach} & \textbf{C/H} & \textbf{F/C} & \textbf{L/B} & \textbf{A/S} & \textbf{P/D} & \textbf{NM} & \textbf{Avg.} & \textbf{Rank}\\
\midrule
Baseline: Frequency MFD &    69.1 &     76.8 &    63.6 &      80.4 &    59.7 &      63.3 &    68.8 &    13.9 \\
unigrams &    86.6 &   \textbf{96.2} &    90.7 &        87.0 &  \textbf{93.1} &      63.2 &    86.1 &    6.1$^{\star\dagger}$ \\
SOTA &      64.0 &       79.0 &      41.0 &        41.0 &      49.0 &       - &    54.8 &    13.4 \\
\hline
\textit{Moral Freq} &    68.2 &     70.7 &    66.1 &      83.6 &    59.8 &        59.0 &    67.9 &    14.1 \\
\textit{Moral Stats} &    79.9 &     80.1 &    78.1 &      92.6 &      74.0 &      57.7 &    77.1 &    10.9 \\
SIMON &    82.2 &     78.7 &      83.0 &      72.5 &    64.8 &        69.0 &      75.0 &    11.9 \\
\hline
SIMON + \textit{Moral Freq} &    81.5 &     82.6 &    80.9 &        74.0 &      68.0 &      69.6 &    76.1 &    10.9 \\
SIMON + \textit{Moral Stats} &    81.5 &     83.4 &    83.7 &      74.2 &    64.9 &      68.9 &    76.1 &      11.0 \\
SIMON + \textit{Moral Freq} + \textit{Moral Stats} &    81.9 &     83.5 &    84.4 &      77.9 &    66.2 &      69.2 &    77.2 &     9.6 \\
\hline
unigrams + \textit{Moral Freq} &    88.4 &     95.8 &    92.3 &      91.8 &    91.1 &      67.3 &    87.8 &    4.0$^\star$ \\
unigrams + \textit{Moral Stats} &    88.1 &     94.4 &  \textbf{92.6} &    \textbf{93.2} &    92.7 &      65.8 &    87.8 &    4.1$^\star$ \\
unigrams + \textit{Moral Freq} + \textit{Moral Stats} &  \textbf{88.5} &     95.1 &    92.5 &    \textbf{93.2} &      93.0 &      66.4 &  \textbf{88.1} &  \textbf{3.1}$^{\star\dagger}$ \\
unigrams + SIMON &    88.1 &     93.6 &    92.3 &      85.9 &    81.4 &      70.9 &    85.4 &    5.6$^{\star\dagger}$ \\
unigrams + SIMON + \textit{Moral Freq} &    88.2 &     91.6 &    91.6 &      86.4 &    80.7 &    \textbf{71.6} &      85.0 &    6.3$^{\star\dagger}$ \\
unigrams + SIMON + \textit{Moral Stats} &    87.4 &     93.1 &    92.1 &      88.1 &    82.7 &      70.1 &    85.6 &    5.9$^{\star\dagger}$ \\
unigrams + SIMON + \textit{Moral Freq} + \textit{Moral Stats} &    87.4 &     93.1 &    91.8 &      88.4 &    83.1 &      70.2 &    85.7 &    5.4$^{\star\dagger}$ \\
\bottomrule
\end{tabular}
\caption{F1-Score of the proposed methods using under-sampling over \textit{Election} \cite{Hoover2019}. C/H: Care/Harm, F/C: Fairness/Cheating, L/B: Loyalty/Betrayal, A/S: Authority/Subversion, P/D: Purity/Degradation, NM: Non-moral, Avg.: Average.`$^\star$' and `$^{\dagger}$' mark that the approach significantly outperforms the MFD baseline and the SOTA, respectively.  baseline. The model with the lowest rank is the one that outperforms the rest.}
\label{tab:election}
\end{table*}

\begin{table*}
\centering
\addtolength{\leftskip}{-3cm}
\addtolength{\rightskip}{-3cm} %extend page margins to center table
\renewcommand{\arraystretch}{1.5}
\begin{tabular}{l|rrrrrr|r}
\toprule
\textbf{Approach} & \textbf{HS} & \textbf{BP} & \textbf{ALM} & \textbf{BLM} & \textbf{D} & \textbf{PE} & \textbf{Rank} \\
\midrule
Baseline: Frequency MFD &      58.0 &      86.1 &    80.5 &    80.5 &       37.0 &     68.8 &      13 \\
unigrams &    83.5 &      67.9 &  \textbf{89.6} &  \textbf{89.6} &   \textbf{81.4} &     86.1 &    4.8$^{\star\dagger}$ \\
SOTA &    51.4 &      86.6 &    65.2 &    65.2 &      4.2 &     54.8 &    13.8 \\
\hline
\textit{Moral Freq} &    59.8 &      78.7 &    80.9 &    80.9 &     38.2 &     67.9 &    14.3 \\
\textit{Moral Stats} &    68.6 &        80.0 &    83.7 &    83.7 &     41.3 &     77.1 &    12.7 \\
SIMON &    75.3 &      80.7 &    83.9 &    83.9 &     78.8 &       75 &    12.3 \\
\hline
SIMON + \textit{Moral Freq} &      76.0 &        81.0 &    84.3 &    84.3 &       79.0 &     76.1 &    11.3 \\
SIMON + \textit{Moral Stats} &    76.8 &      83.1 &    84.6 &    84.6 &     79.4 &     76.1 &     9.9 \\
SIMON + \textit{Moral Freq} + \textit{Moral Stats} &    77.3 &        86.0 &    85.2 &    85.2 &     79.4 &     77.2 &     8.4 \\
\hline
unigrams + \textit{Moral Freq} &    84.3 &      86.6 &    88.8 &    88.8 &     81.2 &     87.8 &  \textbf{2.5}$^{\star\dagger}$ \\
unigrams + \textit{Moral Stats} &    83.4 &    \textbf{86.7} &    86.7 &    86.7 &     81.2 &     87.8 &    4.3$^{\star\dagger}$ \\
unigrams + \textit{Moral Freq} + \textit{Moral Stats} &    83.4 &      82.3 &      87 &      87 &     81.2 &   \textbf{88.1} &    5.3$^{\star\dagger}$ \\
unigrams + SIMON &  \textbf{85.0} &      82.1 &    87.3 &    87.3 &     79.8 &     85.4 &     5.7$^{\dagger}$ \\
unigrams + SIMON + \textit{Moral Freq} &    84.6 &      83.1 &    87.3 &    87.3 &     80.5 &       85 &     5.4$^{\dagger}$ \\
unigrams + SIMON + \textit{Moral Stats} &    84.1 &      83.3 &    86.2 &    86.2 &     80.6 &     85.6 &     6.4 \\
unigrams + SIMON + \textit{Moral Freq} + \textit{Moral Stats} &    84.1 &      83.3 &    86.6 &    86.6 &     80.6 &     85.7 &     5.9$^{\dagger}$ \\
\bottomrule
\end{tabular}
\caption{F1-Score of the proposed methods using under-sampling over all datasets. HS: Hurricane Sandy, BP: Baltimore Protest, ALM: All Lives Matter, BLM: Black Lives Matter, D: Davidson, PE: Presidential Election. `$^\star$' and `$^{\dagger}$' mark that the approach significantly outperforms the MFD baseline and the SOTA, respectively. The model with the lowest rank is the one that outperforms the rest.}
\label{tab:all_datasets}
\end{table*}

\begin{table*}
\centering
\addtolength{\leftskip}{-3cm}
\addtolength{\rightskip}{-3cm} %extend page margins to center table
\renewcommand{\arraystretch}{1.5}
\begin{tabular}{ll|rrrrrr|r}
\toprule
& \textbf{Approach} & \textbf{HS} & \textbf{BP} & \textbf{ALM} & \textbf{BLM} & \textbf{D} & \textbf{PE} & \textbf{Rank} \\
\midrule
\multirow{3}{*}{MFD} & Moral Freq &   58.0 &       59.5 &  68.8 &  80.5 &      37.0 &      68.8 &   5.5 \\
& Moral Stats &   64.8 &       67.3 &  74.1 &  80.2 &      39.2 &      75.6 &   4.0 \\
& SIMON &   74.0 &       \textbf{79.6} &  79.1 &  82.9 &      78.4 &      74.0 &   2.3$^\star$ \\
\hline
\multirow{3}{*}{MoralStrength} & Moral Freq &   59.8 &       61.4 &  68.4 &  80.9 &      38.2 &      67.9 &   5.2 \\
& Moral Stats &   68.6 &       67.9 &  76.6 &  83.7 &      41.3 &      \textbf{77.1} &   2.5$^\star$ \\
& SIMON &   \textbf{75.3} &       78.7 &  \textbf{79.3} &  \textbf{83.9} &      \textbf{78.8} &      75.0 &   \textbf{1.5}$^\star$ \\
\bottomrule
\end{tabular}
\caption{Average F1-Score of the proposed baselines using under-sampling over all datasets. `$^\star$' marks that the approach statistically outperforms the \textit{Moral Freq. with the MFD lexicon}  baseline. The model with the lowest rank is the one that outperforms the rest.}
\label{tab:baselines_comparison}
\end{table*}

\section{Conclusions}
\label{sec:conclusions}

There is an ever-increasing interest in moral values understanding since they reflect our perception, attitudes, and opinion formation on critical societal issues.
Moral values are expressed in user-generated content~\cite{Kalimeri2019}, and primarily through text. With the burst of social media data, emerges a unique opportunity of observing such behaviors in scale and as they happen.
Recent developments in the computational linguistics domain, allow us to analyze automatically such data obtaining useful insights.

Operationalizing morality via the Moral Foundations Theory (MFT)~\cite{Graham2009}, we propose a linguistic resource, \textit{MoralStrength}, that aims at improving the only currently available dictionary, i.e., the MFD.
More specifically,  we contribute with a moral lexicon containing (i) a large number of lemmas, (ii)  less radical and more frequently used lemmas, hence, improving its usability, and (iii) finally, containing a metric of moral valence for each lemma.
\textit{MoralStrength} contains approximately five times more lemmas than the MFD, while at the same time providing with a moral valence, i.e., a quantitative assessment to characterize the lemmas' relationship with each moral dimension.

To explore the potentials of the moral lexicon in predicting the moral narrative in an unseen text, we generated three representations employing a series of feature extraction techniques, including normalized lemmas frequencies, statistical features, and finally, semantic similarity based on word embeddings.
% entailed properties of the 
We evaluated the machine learning framework on six benchmark datasets from the Twitter platform, the only available resources of linguistic data explicitly annotated for their moral content. 
%Having discussed the limitations of the existing resources: the scarcity of annotated datasets, as well as the quality of such annotations.
%The second question tackled the effectiveness of the generated moral-oriented lexicon.

Interestingly, all our models improve the prediction performance with respect to the current state-of-the-art for all moral dimensions. 
The most prominent approaches - as indicated by the Friedman ranking - combine pure textual (e.g., unigrams) with lexicon-based representations (e.g., the \textit{Moral Freq}, the \textit{Moral Stats}, and the \textit{\acs{simon}}).
Hence, we argue that moral lexicon can be successfully employed for moral values classification from a given text since when this information is considered, the models yield higher performance.
%We consider the results obtained from the evaluation on the \textit{Hurricane Sandy}  dataset,  of higher importance, since this the only dataset with annotations obtained by experts.
%Attending to the experimental results, we observe that in all cases this method offers competitive results, and when combined with other features, can yield the best performance in the task at hand.

This study paves the way for further advancements in the moral text analysis, which is indeed an exciting field of study, both from the computational linguistics and the social sciences points of view.
%We find these results compelling, and thus expect that this is explored more in-depth from here on.
From a linguistic perspective, it would be interesting to explore how specific knowledge could be encoded in domain-oriented word vectors, allowing for the development of complex learning methods.
Moreover, the word embedding representation based on moral similarity could be enhanced with the obtained assessments of moral valence, or even combined with sentiment features from the analyzed text.
As for the social sciences, there are numerous issues where detecting the morals narrative can significantly improve our understanding of the peoples' dispositions in, for instance, controversial social phenomena as well as the evolution of opinions over time.

\section*{Acknowledgment}
Oscar Araque has been partially funded by the Spanish Ministry of Economy through the project Semola (TEC2015-68284-R).
Kyriaki Kalimeri acknowledges support from the ``Lagrange Project'' of the ISI Foundation funded by the Fondazione CRT.

%\FloatBarrier

\section*{References}

\bibliographystyle{elsarticle-num}
%\bibliography{./bibtex/bib/IEEEabrv.bib}

\end{document}